\documentclass[final,3p,times]{elsarticle}

\usepackage{amssymb}
\usepackage{amsmath}
\usepackage{multirow}
\usepackage[normalem]{ulem}
\usepackage{booktabs}
\usepackage{subcaption}
\usepackage{bbm}
\usepackage{natbib}
\usepackage[ruled,vlined]{algorithm2e}
\usepackage[colorlinks=true, linkcolor=cyan, filecolor=cyan, urlcolor=cyan, citecolor=cyan]{hyperref}
\usepackage{float}
\usepackage{graphicx}

\useunder{\uline}{\ul}{}

\journal{}

\begin{document}

\begin{frontmatter}

\title{Fine-Grained Representation Learning via Multi-Level Contrastive Learning without Class Priors} 

\author[fafu]{Houwang Jiang}
\ead{hoper.hw@gmail.com}
\author[fafu]{Zhuxian Liu}
\author[fafu]{Guodong Liu}
\author[fafu]{Xiaolong Liu}
\author[fafu]{Shihua Zhan\texorpdfstring{\corref{cor1}}{ (Corresponding author)}}
\ead{zhanshihua2004@fafu.edu.cn}

\cortext[cor1]{Corresponding author.}

\affiliation[fafu]{organization={School of computer science and information engineering},
            addressline={Fujian Agriculture and Forestry University}, 
            city={Fuzhou},
            postcode={350002}, 
            country={China}}

\begin{abstract}
Recent advances in unsupervised representation learning often rely on knowing the number of classes to improve feature extraction and clustering. However, this assumption raises an important question: is the number of classes always necessary, and do class labels fully capture the fine-grained features within the data? In this paper, we propose \textbf{Contrastive Disentangling (CD)}, a framework designed to learn representations without relying on class priors. CD leverages a multi-level contrastive learning strategy, integrating instance-level and feature-level contrastive losses with a normalized entropy loss to capture semantically rich and fine-grained representations. Specifically, (1) the instance-level contrastive loss separates feature representations across samples; (2) the feature-level contrastive loss promotes independence among feature heads; and (3) the normalized entropy loss ensures feature diversity and prevents feature collapse. Extensive experiments on CIFAR-10, CIFAR-100, STL-10, and ImageNet-10 demonstrate that CD outperforms existing methods in scenarios where class information is unavailable or ambiguous. The code is available at \href{https://github.com/Hoper-J/Contrastive-Disentangling}{https://github.com/Hoper-J/Contrastive-Disentangling}.
\end{abstract}



\end{frontmatter}

\section{Introduction}

The growing availability of large-scale unlabeled datasets has underscored the importance of unsupervised learning in extracting meaningful features without predefined class labels. In computer vision, contrastive learning has become a dominant method for unsupervised representation learning, improving feature extraction by maximizing consistency between augmented views of the same instance while minimizing similarity between distinct instances \cite{InvaSpread, SimCLR}.

Although recent contrastive learning methods \cite{SCAN, CC, SeCu} have leveraged class priors to improve performance, their reliance on predefined class counts limits their applicability in real-world scenarios where class information may be uncertain or unavailable. This dependence not only restricts the adaptability of these models but also hampers their ability to generalize to diverse tasks that lack explicit class labels.

To address this limitation, we propose Contrastive Disentangling (CD), a novel unsupervised learning framework that improves feature extraction and disentanglement without requiring class priors.
\begin{figure}[t]
\centering
\includegraphics[width=0.6\textwidth]{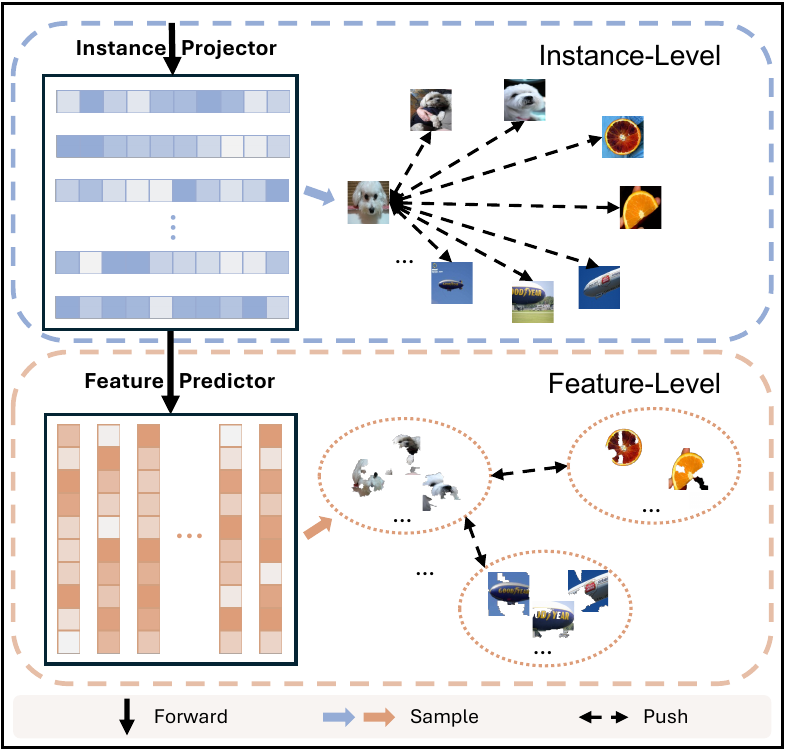}
\caption{\textbf{Overview of the Multi-level Contrastive Learning Strategy in CD Framework.} The instance-level (top) operates on image instances, while the feature-level (bottom) focuses on features visualized using LIME.}
\label{fig:multi-level-contrastive}
\end{figure}
Our approach stems from a detailed observation of the model’s outputs. As illustrated in Figure \ref{fig:multi-level-contrastive}, the rows of the instance projector correspond to the latent representations of the samples, while the columns of the feature predictor represent the predictions of fine-grained features across the entire batch. Traditional contrastive learning methods focus predominantly on instance-level feature representations, yet we raise a new question: can the focus be shifted to the features themselves? Motivated by this insight, we extend contrastive learning to the feature level, treating each feature prediction head as an independent “feature instance” to better capture the underlying attributes of the data. This conceptual shift forms the basis of our CD framework, which integrates multi-level contrastive learning with disentanglement loss to ensure feature diversity and independence, while maintaining a simple and efficient architecture. In summary, the main contributions of this paper are:

\begin{enumerate}
    \item We introduce feature prediction heads and a novel feature-level contrastive approach, enabling the model to capture finer-grained features within a simple and effective framework, providing a fresh and generalizable perspective on contrastive learning.
    
    \item We incorporate a normalized entropy loss to complement feature-level contrastive learning, ensuring that the model captures prevalent and diverse features across the dataset.
    
    \item To the best of our knowledge, this is the first contrastive learning method that achieves semantically rich final outputs without relying on class priors. By defining the model architecture independently of class information, our approach allows researchers to intuitively identify images with similar fine-grained features by examining the final model outputs, which exhibit the same level of semantic richness as the feature extraction layer. Moreover, the model offers researchers the flexibility to adjust the number of feature heads, enabling control over the granularity of the features learned.
\end{enumerate}

Our method demonstrates superior feature extraction capabilities compared to existing unsupervised frameworks and remains highly competitive against methods that rely on class information to constrain model outputs. Additionally, under identical training settings, the final outputs of CD exhibit richer semantics than SimCLR. This improvement is particularly evident in clustering tasks. For example, in the STL-10 dataset, CD achieves a 13.4\% increase in NMI, a 23.1\% increase in ARI, and a 10.9\% improvement in ACC. On the ImageNet-10 dataset, CD shows even greater enhancements, with a 26.8\% increase in NMI, a 110.2\% improvement in ARI, and a 53.1\% increase in ACC.

\section{Related Work}

\subsection{Unsupervised Representation Learning}

Unsupervised representation learning focuses on extracting meaningful features from unlabeled data, with two primary approaches: generative models and contrastive learning. Generative models \cite{AE, VAE, GAN} aim to generate samples that resemble the true data distribution by modeling the underlying latent space. Despite their success in data generation, these models often struggle to learn invariant feature representations—features that remain consistent under various input transformations such as rotation, translation, and scaling \cite{ntelemis2023generic}.

In contrast, instance-level contrastive learning methods \cite{SimCLR, MoCo, BYOL, SimSiam} improve the model’s ability to extract invariant features by maximizing the similarity between augmented views of the same instance while minimizing the similarity between different instances. This strategy has consistently demonstrated strong performance across various downstream tasks, positioning contrastive learning as a highly effective method in unsupervised learning.

\subsection{Utilization of Class Priors}

Recent methods \cite{CC, SeCu, SPICE} incorporate class priors to improve performance by constraining the learning process with known class counts and refining cluster assignments through pseudo-labeling. These approaches have yielded state-of-the-art results in unsupervised clustering, as class priors guide the model towards more discriminative representations.

However, such reliance on class priors significantly limits the adaptability of these models, especially in real-world scenarios where class information is incomplete, ambiguous, or unavailable. This motivates the need for approaches that can effectively capture meaningful and discriminative representations without relying on predefined class information.

\subsection{Multi-level Contrastive Learning}

Multi-level contrastive learning has recently emerged as a novel direction within contrastive learning research. For example, cluster-level contrastive learning by \citet{CC} was introduced for clustering tasks, opening new avenues for structured representation learning. However, methods that partition data based on predefined categories may overlook fine-grained features in more complex datasets, limiting their ability to capture subtle and nuanced patterns.

Building on these observations, our framework extends contrastive learning to both the instance and feature levels. By treating each feature prediction head as an independent “feature instance,” our approach encourages each head to capture distinct, complementary attributes, which is a significant departure from traditional instance-level contrastive learning. This decoupling of feature heads allows CD to learn diverse and fine-grained representations, making it a flexible and powerful method for unsupervised representation learning without relying on class priors.

\section{Method}

\subsection{Proposed Contrastive Disentangling Framework}

\begin{figure*}[htbp]
\centering
\includegraphics[width=0.8\textwidth]{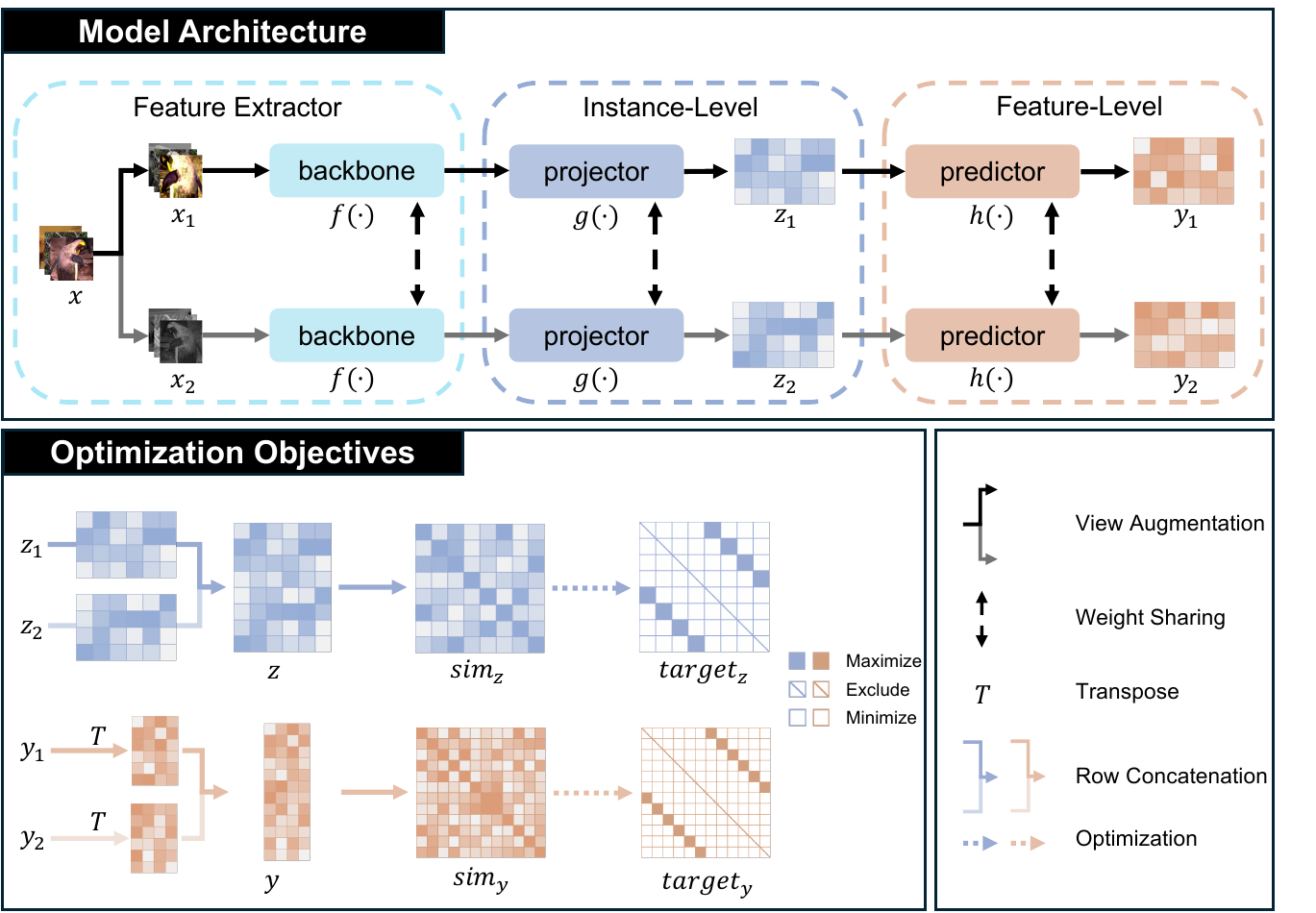}
\caption{\textbf{Architecture of the proposed Contrastive Disentangling (CD) framework.} The upper section illustrates the shared backbone network, instance-level projector, and feature predictor, while the lower section displays the optimization objectives for instance-level and feature-level contrastive learning.}
\label{fig:architecture}
\end{figure*}

The proposed framework employs a multi-level contrastive learning strategy to achieve fine-grained representation learning without the need for class priors, as illustrated in Figure~\ref{fig:architecture}. 
The CD framework comprises the following key components:

\begin{itemize}
    \item \textbf{Data Augmentation}: 
    Each input batch $\mathbf{x}$ undergoes a series of data augmentation operations, including random cropping, horizontal flipping, color jittering, and Gaussian blurring. These augmentations generate two correlated views, denoted as $\mathbf{x}^{(1)}$ and $\mathbf{x}^{(2)}$, of the same sample. This dual-view approach encourages the model to focus on semantically relevant features that are invariant to such transformations.

    \item \textbf{Feature Extraction}: 
    The augmented views are processed by a shared backbone network $f(\cdot)$, implemented using a ResNet architecture \cite{ResNet}. This backbone extracts feature representations $\mathbf{h}^{(1)} = f(\mathbf{x}^{(1)})$ and $\mathbf{h}^{(2)} = f(\mathbf{x}^{(2)})$, ensuring consistent feature extraction across different views.
    
    \item \textbf{Instance Projector}: 
    The extracted features are passed through an instance projector $g(\cdot)$, a multi-layer perceptron (MLP) consisting of two linear layers with batch normalization \cite{BN} and ReLU activation functions. This projector maps the features into a latent space, producing latent representations $\mathbf{z}^{(1)} = g(\mathbf{h}^{(1)})$ and $\mathbf{z}^{(2)} = g(\mathbf{h}^{(2)})$. These representations are then used for instance-level contrastive learning.
    
    \item \textbf{Feature Predictor}: 
    The latent representations are processed by the feature predictor $h(\cdot)$ to capture diverse and disentangled attributes. This module, structurally similar to the instance projector but with a sigmoid activation in the final layer, outputs predictions $\mathbf{y}^{(1)} = h(\mathbf{z}^{(1)})$ and $\mathbf{y}^{(2)} = h(\mathbf{z}^{(2)})$. Each $\mathbf{y}^{(v)}$ is a matrix of size $N \times K$, where $K$ denotes the number of feature heads. Each feature head is designed to capture a distinct attribute of the data.

    \item \textbf{Loss Functions}: 
    The total loss function of CD integrates three components: the instance-level contrastive loss $\mathcal{L}_{\text{inst}}$, the feature-level contrastive loss $\mathcal{L}_{\text{feat}}$, and the normalized entropy loss $\mathcal{L}_{\text{entropy}}$. These losses work together to ensure that the learned features are diverse, independent, and well-distributed, while avoiding feature collapse.

\end{itemize}

\subsection{Instance-level Contrastive Learning}

Instance-level contrastive learning seeks to learn discriminative representations by bringing augmented views of the same instance closer while pushing apart representations of different instances. This dual objective improves the model’s ability to differentiate between instances and capture invariant features across augmentations.

Given an input batch $\mathbf{x}$ of size $N$, two augmented views $\mathbf{x}^{(1)}$ and $\mathbf{x}^{(2)}$ are generated. After passing through the backbone network and instance projector, latent representations $\mathbf{z}^{(1)}$ and $\mathbf{z}^{(2)}$ are obtained. These representations are concatenated to form a combined set of $2N$ latent vectors:
\begin{equation}
\mathbf{z} = \begin{bmatrix} \mathbf{z}^{(1)} \\ \mathbf{z}^{(2)} \end{bmatrix} = \begin{bmatrix} \mathbf{z}_1 \\ \mathbf{z}_2 \\ \vdots \\ \mathbf{z}_{2N} \end{bmatrix} \in \mathbb{R}^{2N \times d},
\label{eq:z_concat}
\end{equation}
where $d$ is the dimensionality of the latent space.

The similarity between any two latent representations is measured using cosine similarity:
\begin{equation}
\text{sim}_z(i, j) = \frac{\mathbf{z}_i \cdot \mathbf{z}_j}{\|\mathbf{z}_i\| \|\mathbf{z}_j\|}
\label{eq:sim_z}
\end{equation}
for $i, j \in \{1, 2, \dots, 2N\}$. 
Cosine similarity measures the relationship between latent representations, with higher values indicating greater similarity. Building on this similarity measure, we define the instance-level contrastive loss $\mathcal{L}_{\text{inst}}$, which aims to maximize the similarity between positive pairs (i.e., the other augmented view of the same instance) and minimize the similarity between negative pairs (i.e., views of different instances). For each sample $i$, the instance-level contrastive loss is formulated as:
\begin{equation}
\ell_i^{\text{inst}} = -\log \frac{\exp(\text{sim}_z(i, \text{pos}(i)) / \tau_{\text{inst}})}{\sum_{k=1}^{2N} \mathbbm{1}_{[k \neq i]} \exp(\text{sim}_z(i, k) / \tau_{\text{inst}})}
\label{eq:inst_loss_i}
\end{equation}
where $\text{pos}(i)$ denotes the index of the positive sample corresponding to $i$, $\tau_{\text{inst}}$ is a temperature parameter controlling the concentration level of the distribution, and $\mathbbm{1}_{[k \neq i]}$ is an indicator function that equals 1 when $k \neq i$ and 0 otherwise. The total instance-level contrastive loss is computed by averaging over all $2N$ samples:
\begin{equation}
\mathcal{L}_\text{inst} = \frac{1}{2N} \sum_{i=1}^{2N} \ell_i^{\text{inst}}
\label{eq:inst_loss}
\end{equation}

\subsection{Disentanglement Loss}

To ensure that the predicted feature attributes capture meaningful variation across the dataset, we introduce two complementary loss functions: the feature-level contrastive loss, which promotes independence by encouraging each feature head to focus on distinct attributes, and the normalized entropy loss, which ensures that the learned features are well-distributed and representative of the overall data distribution.

\subsubsection{Feature-Level Contrastive Loss}

The feature predictor $h(\cdot)$ generates predictions $\mathbf{y}^{(1)}$ and $\mathbf{y}^{(2)}$, each of size $N \times K$, where $K$ is the number of feature heads. In our framework, each feature head is treated as a distinct “feature instance” that captures different characteristics of the dataset. By introducing feature-level contrastive learning, our method uniquely enables each head to capture complementary and independent attributes. To maintain consistency in the formulation, we transpose the output matrix of the feature heads and concatenate the outputs from both views:
\begin{equation}
\mathbf{y} = \begin{bmatrix} (\mathbf{y}^{(1)})^\top \\ (\mathbf{y}^{(2)})^\top \end{bmatrix} = \begin{bmatrix} \mathbf{y}_1 \\ \mathbf{y}_2 \\ \vdots \\ \mathbf{y}_{2K} \end{bmatrix} \in \mathbb{R}^{2K \times N},
\label{eq:y_concat}
\end{equation}
where $\mathbf{y}_i$ represents the predictions of feature head $i$ across the batch. The cosine similarity between any two feature prediction vectors is defined as:
\begin{equation}
\text{sim}_y(i, j) = \frac{\mathbf{y}_i \cdot \mathbf{y}_j}{\|\mathbf{y}_i\| \|\mathbf{y}_j\|}
\label{eq:sim_y}
\end{equation}

The feature-level contrastive loss promotes the independence of feature heads by maximizing the similarity between corresponding heads across different views while minimizing the similarity between different heads. For feature head $i$, the loss is defined as:
\begin{equation}
\ell_i^{\text{feat}} = -\log \frac{\exp(\text{sim}_y(i, \text{pos}(i)) / \tau_{\text{feat}})}{\sum_{k=1}^{2K} \mathbbm{1}_{[k \neq i]} \exp(\text{sim}_y(i, k) / \tau_{\text{feat}})}
\label{eq:feat_loss_i}
\end{equation}
where $\text{pos}(i)$ denotes the index of the positive sample for feature head $i$, and $\tau_{\text{feat}}$ is the temperature parameter for the feature-level contrastive loss. The total feature-level contrastive loss is computed by averaging over all $2K$ feature heads:
\begin{equation}
\mathcal{L}_\text{feat} = \frac{1}{2K} \sum_{i=1}^{2K} \ell_i^{\text{feat}}
\label{eq:feat_loss}
\end{equation}

\subsubsection{Normalized Entropy Loss}

To prevent feature collapse and ensure that the learned features are meaningful and widespread across the dataset, we introduce a normalized entropy loss $\mathcal{L}_{\text{entropy}}$. This loss encourages each feature head to distribute its predictions evenly, capturing non-trivial and prevalent patterns in the data.
The normalized binary entropy for each feature prediction vector $\mathbf{y}_i$ is defined as:
\begin{equation}
H(\mathbf{y}_i) = -\frac{1}{N \log(2)} \sum_{j=1}^{N} \left( \mathbf{y}_{ij} \log(\mathbf{y}_{ij}) + (1 - \mathbf{y}_{ij}) \log(1 - \mathbf{y}_{ij}) \right)
\label{eq:entropy_loss_i}
\end{equation}

The overall normalized entropy loss is the average entropy across all feature heads:

\begin{equation}
\mathcal{L}_\text{entropy} = \frac{1}{2K} \sum_{i=1}^{2K} H(\mathbf{y}_i)
\label{eq:entropy_loss}
\end{equation}

\subsubsection{Combined Disentanglement Loss}

The total disentanglement loss is the combination of the feature-level contrastive loss and the normalized entropy loss:
\begin{equation}
\mathcal{L}_{\text{disentanglement}} = \mathcal{L}_{\text{feat}} - \alpha \mathcal{L}_{\text{entropy}}
\label{eq:disentanglement_loss}
\end{equation}
where $\alpha$ is a hyperparameter that controls the balance between encouraging diversity and ensuring meaningful feature representations. By simultaneously optimizing these two objectives, the feature heads are encouraged to capture both diverse patterns and informative representations, leading to better feature disentanglement.

\subsection{Total Loss}

The total loss function for training the CD framework is the sum of the instance-level contrastive loss and the disentanglement loss. This ensures that both the instance-level discrimination and feature-level disentanglement are optimized simultaneously:
\begin{equation}
\mathcal{L}_{\text{total}} = \mathcal{L}_{\text{inst}} + \mathcal{L}_{\text{disentanglement}}.
\label{eq:total_loss}
\end{equation}

By jointly optimizing instance and feature-level objectives, our framework achieves a unique balance between learning discriminative representations and ensuring feature diversity, all without relying on class priors. This makes our approach especially suited for complex unsupervised learning tasks where class information is scarce or completely unavailable, providing a robust solution for learning fine-grained and diverse representations.
Algorithm \ref{alg:CD} summarizes the proposed method.

\begin{algorithm}[H]
\caption{Contrastive Disentangling Framework}
\label{alg:CD}
\KwIn{Dataset $X$, Batch size $N$, Feature number $K$, Temperature constants $\tau_{\text{inst}}$ and $\tau_{\text{feat}}$}
\KwOut{Total loss $\mathcal{L}_{\text{total}}$}
\SetAlgoLined

\For{each minibatch $\mathbf{x}$ of size $N$ in $X$}{
    
    \tcp{Apply data augmentation}
    $\mathbf{x} \rightarrow \{\mathbf{x}^{(1)}, \mathbf{x}^{(2)}\}$\;
    
    \For{each view $\mathbf{x}^{(v)}, v \in \{1, 2\}$}{
        \tcp{Extract features}
        $\mathbf{h}^{(v)} = f(\mathbf{x}^{(v)})$\;
        
        \tcp{Project to latent space}
        $\mathbf{z}^{(v)} = g(\mathbf{h}^{(v)})$\;
        
        \tcp{Generate feature head predictions}
        $\mathbf{y}^{(v)} = h(\mathbf{z}^{(v)})$\;
    }

    Combine $\mathbf{z^{(1)}}$, $\mathbf{z^{(2)}}$ by Eq.(\ref{eq:z_concat}) and $\mathbf{y^{(1)}}$, $\mathbf{y^{(2)}}$ by Eq.(\ref{eq:y_concat})

    \tcp{Compute Losses}
    Compute cosine similarities $\text{sim}_z(i, j) $,  $\text{sim}_y(i, j)$ by Eq.~(\ref{eq:sim_z}), (\ref{eq:sim_y});

    Compute instance-level contrastive loss $\mathcal{L}_\text{inst}$ by Eq.~(\ref{eq:inst_loss_i}), (\ref{eq:inst_loss});
    
    Compute feature-level contrastive loss $\mathcal{L}_\text{feat}$ by Eq.~(\ref{eq:feat_loss_i}), (\ref{eq:feat_loss});
    
    Compute normalized entropy loss $\mathcal{L}_\text{entropy}$ by Eq.~(\ref{eq:entropy_loss_i}), (\ref{eq:entropy_loss});

    Compute total loss $\mathcal{L}_{\text{total}}$ by Eq.~(\ref{eq:total_loss});

    \tcp{Update Model Parameters}
    Update $f$, $g$, and $h$ by minimizing $\mathcal{L}_{\text{total}}$;
}
\Return $\mathcal{L}_{\text{total}}$
\end{algorithm}

\section{Experiment}

\subsection{Datasets}

We conducted experiments on four widely-used datasets: CIFAR-10, CIFAR-100 \cite{CIFAR}, STL-10 \cite{STL10}, and ImageNet-10 \cite{ImageNet}. These datasets serve as benchmarks to rigorously evaluate the effectiveness of our approach. Table \ref{tab:datasets} provides an overview of the dataset splits, sample sizes, and the number of classes.

\begin{table}[h]
  \centering
  \caption{Overview of datasets used for training and evaluation}
    \begin{tabular}{lccc}
    \toprule
    Dataset & Split & Samples & Classes \\
    \midrule
    CIFAR-10 & train+test & 60,000 & 10 \\
    CIFAR-100-20 & train+test & 60,000 & 20 \\
    STL-10 & train+test & 13,000 & 10 \\
    ImageNet-10 & train & 13,000 & 10 \\
    \bottomrule
    \end{tabular}%
  \label{tab:datasets}%
\end{table}%

For CIFAR-100, we used 20 super-classes as ground-truth labels instead of the 100 fine-grained classes, aligning with benchmark settings in previous studies. Additionally, to guarantee the rigor of our assessment, no unseen data (e.g., the unlabeled data in STL-10) was used during training.

\subsection{Experimental Settings}

\textbf{Data Augmentation.} 
To ensure comparability with existing methods \cite{CC}, we adopted the same data augmentation strategies, using ResNet-34 as the backbone for feature extraction. All datasets were subject to random cropping to 224x224, horizontal flipping, color jittering, and grayscale transformation. Additionally, Gaussian blur was applied to the ImageNet-10 dataset, while the upscaling of smaller datasets (CIFAR-10, CIFAR-100, and STL-10) provided a comparable blurring effect.

\textbf{Training Details.} 
In our experiments, both the hidden dimension and the number of features were set equal to the batch size. A cosine decay schedule without restarts \cite{SimCLR, SGDR} and gradient clipping were employed to stabilize training, as validated in our ablation studies (Table \ref{tab:sch&clip}). No additional techniques, such as self-labeling, were used to improve performance. The temperature parameters $\tau_{\text{inst}}$ and $\tau_{\text{feat}}$ were set to 0.5 and 1, respectively, while the balance parameter $\alpha$ for feature diversity was set to 1. Other training settings were based on prior contrastive learning works, including 1000 training epochs, a random seed of 42, and the Adam optimizer \cite{Adam} with a learning rate of $3 \times 10^{-4}$ and no weight decay.

\textbf{Environment.} 
The experiments were performed on an NVIDIA RTX 3090 GPU (24GB) and an 18 vCPU AMD EPYC 9754 128-Core Processor. Batch sizes of 128 and 256 were used. Training time for CIFAR-10 and CIFAR-100 was approximately 48 GPU hours, while for STL-10 and ImageNet-10, it was around 11 GPU hours. Detailed logs of all experimental procedures are available \href{https://github.com/Hoper-J/Contrastive-Disentangling/tree/master?tab=readme-ov-file#experiment-records}{here}.

\textbf{Evaluation Metrics.} 
We used three clustering metrics — Normalized Mutual Information (NMI), Adjusted Rand Index (ARI), and Accuracy (ACC) — to evaluate the feature extraction and disentanglement capabilities of the model.

\subsection{Compared Methods}

We compared CD against a diverse set of representative unsupervised models, selected based on their relevance to the challenges addressed by our approach. In addition to the traditional K-means clustering algorithm \cite{kmeans}, we assessed fully unsupervised models such as AE \cite{AE}, DeCNN \cite{DeCNN}, DAE \cite{DAE}, VAE \cite{VAE}, and DCGAN \cite{DCGAN}, all of which do not rely on class information during training, like our CD model.

Furthermore, we benchmarked CD against unsupervised models that incorporate class number information, including JULE \cite{JULE}, DEC \cite{DEC}, DAC \cite{DAC}, DCCM \cite{DCCM}, PICA \cite{PICA}, and CC \cite{CC}. All experiments were conducted using the full dataset and adhered to identical training and evaluation protocols to ensure fairness.

\subsection{Experimental Results}

We evaluated CD at two key stages: feature extraction and feature disentanglement. K-means clustering was applied to both the backbone output and the final model output to assess performance.

\textbf{Backbone.} 
Table \ref{tab:backbone} presents a comparative analysis at the feature extraction stage using NMI, ARI, and ACC metrics. Across all datasets — CIFAR-10, CIFAR-100, STL-10, and ImageNet-10 — both variants of CD (CD-128 and CD-256) outperformed fully unsupervised methods such as AE, DeCNN, DAE, VAE, and DCGAN. Notably, CD achieved the highest NMI scores on all datasets, with significant improvements on STL-10 and ImageNet-10, and superior overall performance compared to the class-dependent model CC.

\begin{table}[t]
\centering
    \caption{\textbf{Feature extraction performance comparison across models.} The 1\textsuperscript{st}/2\textsuperscript{nd} best results are indicated in \textbf{bold}/\uline{underlined}. Models that utilized class number information during training are marked with *.}
    \resizebox{\linewidth}{!}{
    \begin{tabular}{lcccccccccccc}
    \toprule
    \multicolumn{1}{l}{Dataset} & \multicolumn{3}{l}{CIFAR-10}                                                & \multicolumn{3}{l}{CIFAR-100-20}                                            & \multicolumn{3}{l}{STL-10}                                                  & \multicolumn{3}{l}{ImageNet-10}                                             \\ \hline
    \multicolumn{1}{l}{Metrics} & \multicolumn{1}{l}{NMI} & \multicolumn{1}{l}{ARI} & \multicolumn{1}{l}{ACC} & \multicolumn{1}{l}{NMI} & \multicolumn{1}{l}{ARI} & \multicolumn{1}{l}{ACC} & \multicolumn{1}{l}{NMI} & \multicolumn{1}{l}{ARI} & \multicolumn{1}{l}{ACC} & \multicolumn{1}{l}{NMI} & \multicolumn{1}{l}{ARI} & \multicolumn{1}{l}{ACC} \\ \hline
    K-means                     & 0.087                   & 0.049                   & 0.229                   & 0.084                   & 0.028                   & 0.130                   & 0.125                   & 0.061                   & 0.192                   & 0.119                   & 0.057                   & 0.241                   \\
    AE                          & 0.239                   & 0.169                   & 0.314                   & 0.100                   & 0.048                   & 0.165                   & 0.250                   & 0.161                   & 0.303                   & 0.210                   & 0.152                   & 0.317                   \\
    DeCNN                       & 0.240                   & 0.174                   & 0.282                   & 0.092                   & 0.038                   & 0.133                   & 0.227                   & 0.162                   & 0.299                   & 0.186                   & 0.142                   & 0.313                   \\
    DAE                         & 0.251                   & 0.163                   & 0.297                   & 0.111                   & 0.046                   & 0.151                   & 0.224                   & 0.152                   & 0.302                   & 0.206                   & 0.138                   & 0.304                   \\
    VAE                         & 0.245                   & 0.167                   & 0.291                   & 0.108                   & 0.040                   & 0.152                   & 0.200                   & 0.146                   & 0.282                   & 0.193                   & 0.168                   & 0.334                   \\
    GAN                         & 0.265                   & 0.176                   & 0.315                   & 0.120                   & 0.045                   & 0.151                   & 0.210                   & 0.139                   & 0.298                   & 0.225                   & 0.157                   & 0.346                   \\
    DCGAN                       & 0.265                   & 0.176                   & 0.315                   & 0.120                   & 0.045                   & 0.151                   & 0.210                   & 0.139                   & 0.298                   & 0.225                   & 0.157                   & 0.346                   \\
    CC*                          & 0.670                   & 0.523                   & 0.684                   & 0.433                   & \textbf{0.244}          & \textbf{0.431}          & 0.608                   & 0.443                   & 0.594                   & 0.800                   & 0.695                   & 0.846                   \\ \hline
    CD-128 (Ours)                      & {\ul 0.725}             & {\ul 0.620}             & {\ul 0.800}             & {\ul 0.462}             & {\ul 0.240}             & 0.418                   & {\ul 0.670}             & {\ul 0.523}             & {\ul 0.684}             & \textbf{0.893}          & \textbf{0.858}          & {\ul 0.927}             \\
    CD-256 (Ours)                      & \textbf{0.734}          & \textbf{0.635}          & \textbf{0.807}          & \textbf{0.476}          & 0.231                   & {\ul 0.422}             & \textbf{0.687}          & \textbf{0.581}          & \textbf{0.758}          & {\ul 0.885}             & {\ul 0.854}             & \textbf{0.934}          \\ \bottomrule
    \end{tabular}
    }
\label{tab:backbone}
\end{table}

\textbf{Final output.} 
Table \ref{tab:feature} compares CD to models that use class number information. CD demonstrated strong performance across all metrics, particularly on more challenging datasets like STL-10 and ImageNet-10. While ARI and ACC were slightly lower on CIFAR-10 and CIFAR-100 compared to CC, this is primarily due to the lower complexity of these smaller datasets, which limits the advantage of CD’s ability to disentangle fine-grained features. In datasets with more intricate patterns and greater feature diversity, CD’s strengths are more effectively utilized. Importantly, CD achieved an ACC of 0.734 with a batch size of 256, surpassing PICA$^\dagger$ (0.713) despite not using additional unlabeled data, which consists of 100,000 samples.

\begin{table}[t]
\caption{\textbf{Final output performance comparison across models.} The 1\textsuperscript{st}/2\textsuperscript{nd} best results are indicated in \textbf{bold}/\uline{underlined}. Models that utilized class number information during training are marked with *, while those that utilized unlabeled datasets for STL-10 are marked with $^\dagger$.}
\resizebox{\linewidth}{!}{
    \begin{tabular}{lllllllllllll}
    \toprule
    Dataset & \multicolumn{3}{l}{CIFAR-10}                     & \multicolumn{3}{l}{CIFAR-100-20}                 & \multicolumn{3}{l}{STL-10}                       & \multicolumn{3}{l}{ImageNet-10}                  \\ \hline
    Metrics & NMI            & ARI            & ACC            & NMI            & ARI            & ACC            & NMI            & ARI            & ACC            & NMI            & ARI            & ACC            \\ \hline
    JULE*    & 0.192          & 0.138          & 0.272          & 0.103          & 0.033          & 0.137          & 0.182          & 0.164          & 0.277          & 0.175          & 0.138          & 0.300          \\
    DEC*     & 0.257          & 0.161          & 0.301          & 0.136          & 0.050          & 0.185          & 0.276          & 0.186          & 0.359          & 0.282          & 0.203          & 0.381          \\
    DAC*$^\dagger$     & 0.396          & 0.306          & 0.522          & 0.185          & 0.088          & 0.238          & 0.366          & 0.257          & 0.470          & 0.394          & 0.302          & 0.527          \\
    DCCM*    & 0.496          & 0.408          & 0.623          & 0.285          & 0.173          & 0.327          & 0.376          & 0.262          & 0.482          & 0.608          & 0.555          & 0.710          \\
    PICA*$^\dagger$    & 0.591          & 0.512          & 0.696          & 0.310          & 0.171          & 0.337          & 0.611          & 0.531          & 0.713          & 0.802          & 0.761          & 0.870          \\
    CC*      & 0.705          & \textbf{0.637} & \textbf{0.790} & 0.417          & 0.221          & \textbf{0.421} & 0.622          & 0.539          & 0.670          & 0.859          & 0.822          & 0.893          \\ \hline
    CD-128 (Ours)  & \textbf{0.711} & {\ul 0.624}    & {\ul 0.788}    & {\ul 0.438}    & {\ul 0.249}    & 0.394          & \textbf{0.687} & {\ul 0.549}    & {\ul 0.702}    & \textbf{0.898} & \textbf{0.869} & \textbf{0.932} \\
    CD-256 (Ours)  & {\ul 0.706}    & 0.621          & 0.782          & \textbf{0.446} & \textbf{0.254} & {\ul 0.416}    & {\ul 0.668}    & \textbf{0.572} & \textbf{0.734} & {\ul 0.887}    & {\ul 0.861}    & {\ul 0.928}    \\ \bottomrule
    \end{tabular}
}
\label{tab:feature}
\end{table}

These results strongly support the effectiveness of CD in achieving high performance without relying on class priors. Furthermore, CD demonstrated excellent performance even with smaller batch sizes. For a t-SNE visualization of the CD-256 model’s feature representations, refer to \ref{app:tsne}.

\subsection{Ablation Studies}
\label{subsec:ablation_study}

We conducted ablation studies on STL-10 and ImageNet-10 with a batch size of 128 to assess the contributions of different components of our model. In Sections \ref{subsec:ne&bn} and \ref{subsec:tricks}, we incrementally added modules to validate their effectiveness, while in Sections \ref{subsec:featurehead} and \ref{subsec:aug}, we removed certain modules to verify their necessity.

\subsubsection{Normalized Entropy Loss and Batch Normalization}
\label{subsec:ne&bn}

\begin{figure}[t]
    \centering
    \begin{subfigure}[b]{0.45\linewidth}
        \centering
        \includegraphics[width=\linewidth]{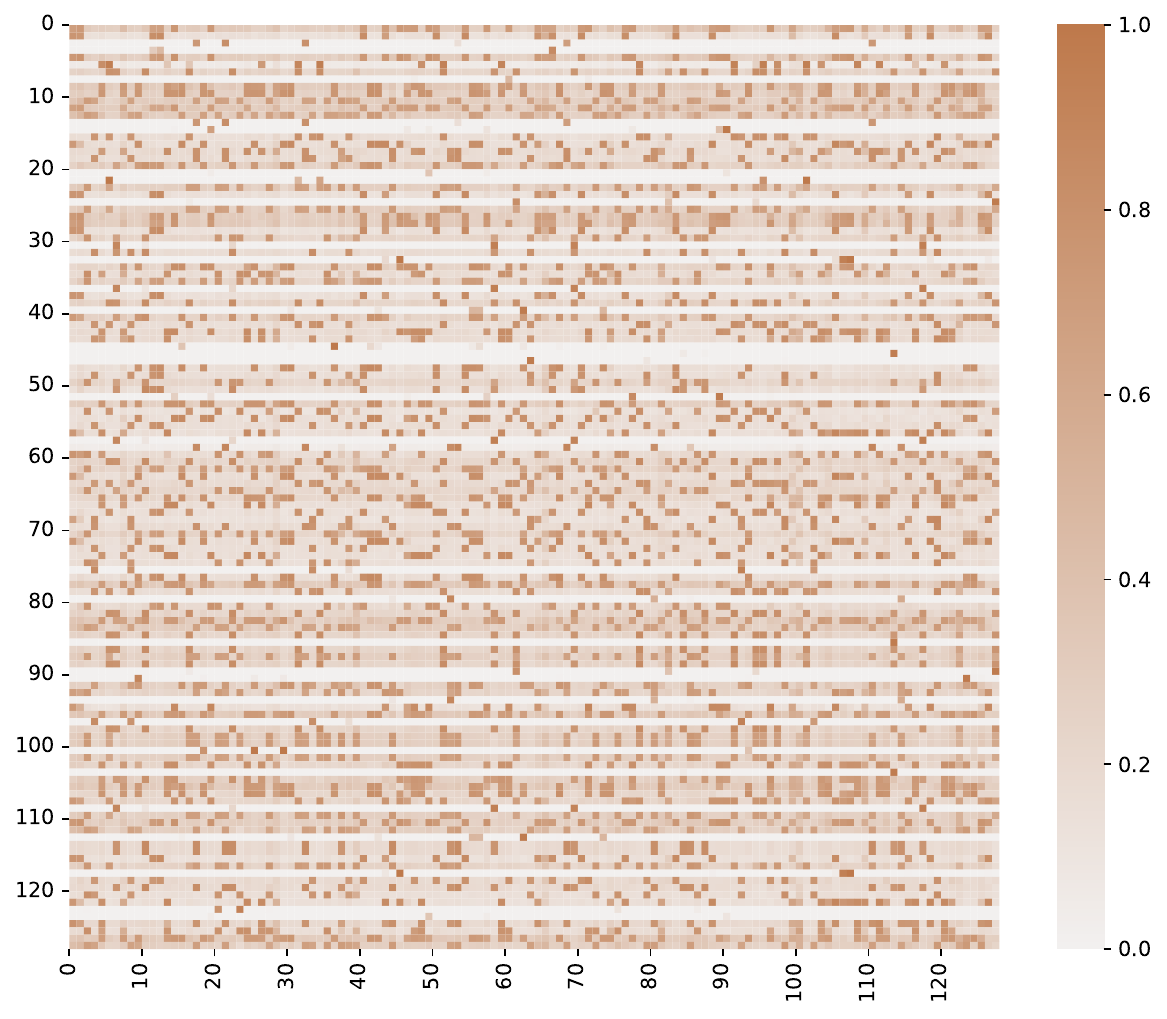}
        \caption{NE loss}
        \label{fig:default}
    \end{subfigure}
    \hfill
    \begin{subfigure}[b]{0.45\linewidth}
        \centering
        \includegraphics[width=\linewidth]{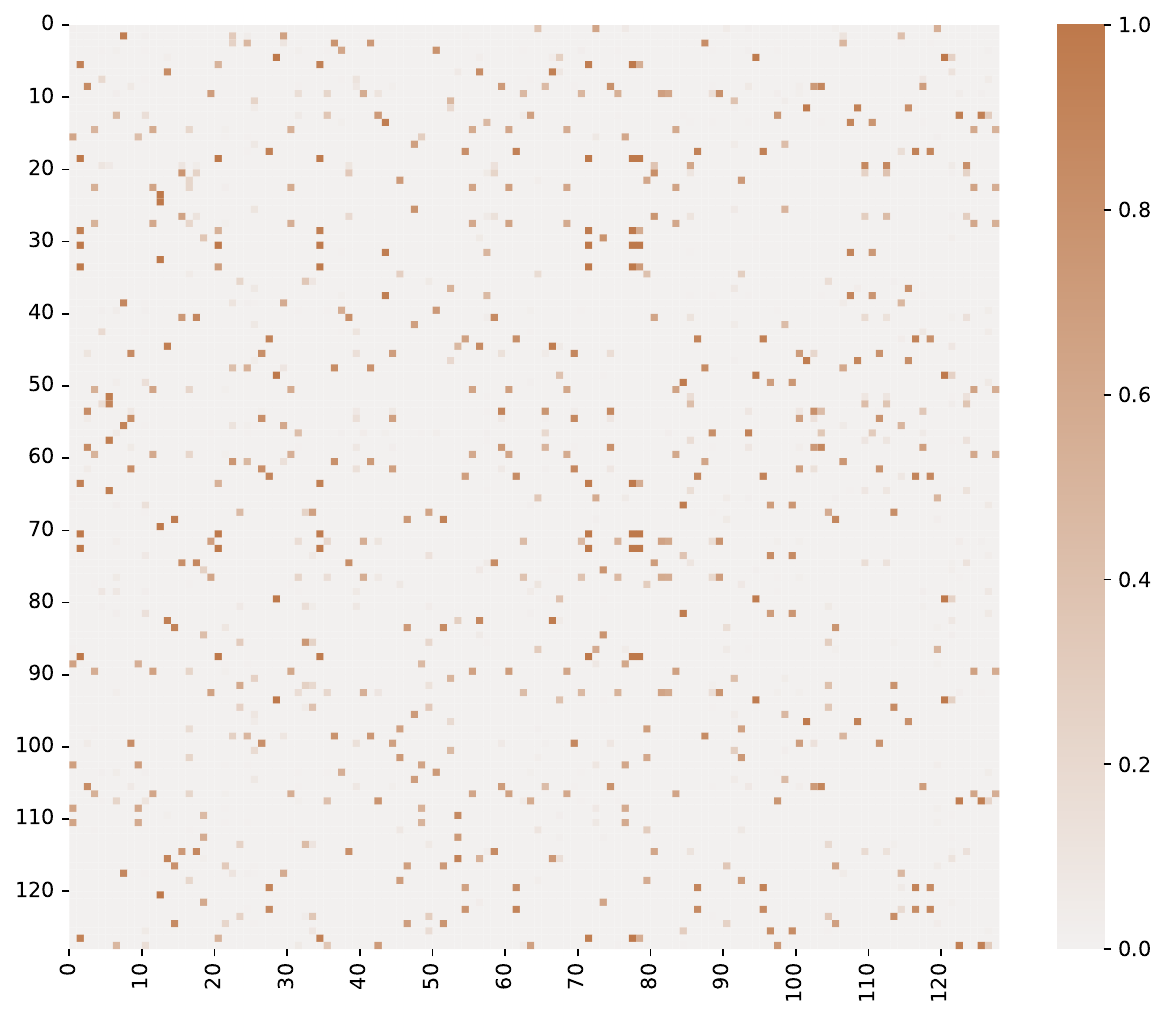}
        \caption{no NE loss}
        \label{fig:noneloss}
    \end{subfigure}
    \caption{\textbf{Visualization of transposed model outputs with/without normalized entropy loss (NE loss)}. The left figure (a) shows outputs with NE loss, emphasizing common features, while the right figure (b) shows outputs without NE loss, capturing more image-specific features.}
    \label{fig:ablation neloss}
\end{figure}

Table \ref{tab:NE&BN} demonstrates the impact of the normalized entropy loss (NE) and batch normalization (BN) on performance. Both components significantly improved results, with further enhancements observed when combined. 
\begin{table}[t]
\centering
  \caption{Effect of normalized entropy loss (NE) and batch normalization (BN) on performance across backbone and final output.}
  \resizebox{\linewidth}{!}{
    \begin{tabular}{rllccclccc}
    \toprule
    \multicolumn{1}{l}{Dataset}     & NE                        & BN                        & \multicolumn{3}{l}{Backbone}                                                         &  & \multicolumn{3}{l}{Feature}                                                          \\ \cline{4-6} \cline{8-10} 
                                    &                           &                           & \multicolumn{1}{l}{NMI}    & \multicolumn{1}{l}{ARI}    & \multicolumn{1}{l}{ACC}    &  & \multicolumn{1}{l}{NMI}    & \multicolumn{1}{l}{ARI}    & \multicolumn{1}{l}{ACC}    \\ \hline
    \multicolumn{1}{l}{STL-10}      & -                         & -                         & $0.519 \pm 0.029$          & $0.265 \pm 0.037$          & $0.538 \pm 0.050$          &  & $0.352 \pm 0.040$          & $0.065 \pm 0.028$          & $0.354 \pm 0.041$          \\
                                    & \checkmark & -                         & $0.638 \pm 0.021$          & $0.482 \pm 0.035$          & $0.695 \pm 0.031$          &  & $0.656 \pm 0.005$          & $0.527 \pm 0.004$          & $0.671 \pm 0.007$          \\
                                    & -                         & \checkmark & $0.659 \pm 0.017$          & $0.495 \pm 0.036$          & $0.653 \pm 0.037$          &  & $0.367 \pm 0.052$          & $0.069 \pm 0.034$          & $0.306 \pm 0.044$          \\
                                    & \checkmark & \checkmark & $\mathbf{0.677 \pm 0.020}$ & $\mathbf{0.547 \pm 0.031}$ & $\mathbf{0.705 \pm 0.039}$ &  &  $\mathbf{0.669 \pm 0.006}$ & $\mathbf{0.543 \pm 0.016}$ & $\mathbf{0.681 \pm 0.020}$ \\ \hline
    \multicolumn{1}{l}{ImageNet-10} & -                         & -                         & $0.632 \pm 0.022$          & $0.287 \pm 0.041$          & $0.620 \pm 0.042$          &  & $0.549 \pm 0.038$          & $0.163 \pm 0.037$          & $0.424 \pm 0.040$          \\
                                    & \checkmark & -                         & $0.748 \pm 0.020$          & $0.532 \pm 0.054$          & $0.779 \pm 0.047$          &  & $0.887 \pm 0.002$          & $0.861 \pm 0.002$          & $0.927 \pm 0.001$          \\
                                    & -                         & \checkmark & $0.837 \pm 0.018$          & $0.733 \pm 0.050$          & $0.831 \pm 0.055$          &  & $0.538 \pm 0.043$          & $0.152 \pm 0.045$          & $0.416 \pm 0.055$          \\
                                    & \checkmark & \checkmark & $\mathbf{0.857 \pm 0.021}$ & $\mathbf{0.772 \pm 0.068}$ & $\mathbf{0.858 \pm 0.071}$ &  & $\mathbf{0.890 \pm 0.004}$ & $\mathbf{0.864 \pm 0.006}$ & $\mathbf{0.929 \pm 0.004}$ \\ \bottomrule
        \end{tabular}
    }
\label{tab:NE&BN}
\end{table}
To illustrate the effect of NE loss on feature disentanglement, Figure \ref{fig:ablation neloss} visualizes the model's outputs with and without NE loss. This visualization highlights the importance of NE loss in guiding the model to capture common features across the dataset.

The rows in Figure \ref{fig:ablation neloss} represent the feature attributes learned by our model, as depicted in the feature-level part of Figure \ref{fig:multi-level-contrastive}. Further visualization of these attributes using LIME is available in \ref{app:lime}.

\subsubsection{Training Tricks}
\label{subsec:tricks}

Next, we evaluated the contribution of cosine decay scheduling and gradient clipping (Table \ref{tab:sch&clip}). The results show that both techniques enhanced performance and stabilized training.

\begin{table*}[t]
    \centering
    \caption{Effect of cosine decay schedule (Sch.) and gradient clipping (Clip.) on model performance.}
        \begin{tabular}{lllll}
        \toprule
        Dataset     & Sch. \& Clip. & NMI                    & ARI                    & ACC                    \\ \hline
        STL-10      & -             & $0.669 \pm 0.006$      & $0.543 \pm 0.016$      & $0.681 \pm 0.020$      \\
                    & \checkmark    & $\mathbf{0.679 \pm 0.007}$ & $\mathbf{0.564 \pm 0.023}$ & $\mathbf{0.720 \pm 0.026}$ \\ \hline
        ImageNet-10 & -             & $0.890 \pm 0.004$      & $0.864 \pm 0.006$      & $0.929 \pm 0.004$      \\
                    & \checkmark    & $\mathbf{0.898 \pm 0.003}$ & $\mathbf{0.868 \pm 0.006}$ & $\mathbf{0.931 \pm 0.003}$ \\    \bottomrule 
        \end{tabular}
  \label{tab:sch&clip}
\end{table*}

\subsubsection{Feature Prediction Head}
\label{subsec:featurehead}

We evaluated the effect of removing the feature prediction head, which reduces the model to an instance-level contrastive learning framework similar to SimCLR. As shown in Table \ref{tab:feature-level}, the feature prediction head significantly boosts performance, and its removal results in substantial degradation. This highlights the critical role of feature-level contrastive learning in effective disentanglement (see Table \ref{tab:appendix_featurelevel} for backbone results). These findings demonstrate that our feature-level contrastive learning approach enables the model to capture more distinct and diverse semantic features, resulting in improved representation quality in the final output.

\begin{table*}[t]
  \centering
  \caption{Effect of feature-level head on model performance.}
    \begin{tabular}{rlccc}
    \toprule
    \multicolumn{1}{l}{Dataset} & Feature-Level Head & \multicolumn{1}{l}{NMI} & \multicolumn{1}{l}{ARI} & \multicolumn{1}{l}{ACC} \\
    \midrule
    \multicolumn{1}{l}{STL-10} & -     & $0.599 \pm 0.029$ & $0.458 \pm 0.043$ & $0.649 \pm 0.053$ \\
          & \checkmark     & $\mathbf{0.679 \pm 0.007}$ & $\mathbf{0.564 \pm 0.023}$ & $\mathbf{0.720 \pm 0.026}$ \\
    \midrule
    \multicolumn{1}{l}{ImageNet-10} & -     & $0.708 \pm 0.031$ & $0.413 \pm 0.060$ & $0.608 \pm 0.065$ \\
          & \checkmark     & $\mathbf{0.898 \pm 0.003}$ & $\mathbf{0.868 \pm 0.006}$ & $\mathbf{0.931 \pm 0.003}$ \\
    \bottomrule
    \end{tabular}%
  \label{tab:feature-level}%
\end{table*}%

\subsubsection{Data Augmentation}
\label{subsec:aug}

Finally, we examined the necessity of the dual-view data augmentation strategy (Table \ref{tab:augmentation}). The results indicate that dual-view augmentation is critical for feature disentanglement, whereas single-view augmentation can still achieve satisfactory performance during feature extraction (see Table \ref{tab:appendix_augmentation} for details).

\begin{table*}[ht]
  \centering
  \caption{Effect of single-view vs. dual-view data augmentation on model performance.}
    \begin{tabular}{rllccc}
    \toprule
    \multicolumn{1}{l}{Dataset} & $x^{(1)}$    & $x^{(2)}$    & \multicolumn{1}{l}{NMI} & \multicolumn{1}{l}{ARI} & \multicolumn{1}{l}{ACC} \\
    \midrule
    \multicolumn{1}{l}{STL-10} & -     & \checkmark     & $0.643 \pm 0.004$ & $0.555 \pm 0.012$ & $0.717 \pm 0.014$ \\
          & \checkmark     & \checkmark     & $\mathbf{0.679 \pm 0.008}$ & $\mathbf{0.564 \pm 0.023}$ & $\mathbf{0.720 \pm 0.026}$ \\
    \midrule
    \multicolumn{1}{l}{ImageNet-10} & -     & \checkmark     & $0.874 \pm 0.001$ & $0.847 \pm 0.001$ & $0.918 \pm 0.001$ \\
          & \checkmark     & \checkmark     & $\mathbf{0.898 \pm 0.003}$ & $\mathbf{0.868 \pm 0.006}$ & $\mathbf{0.931 \pm 0.003}$ \\
    \bottomrule
    \end{tabular}%
  \label{tab:augmentation}%
\end{table*}%

\section{Conclusion}

In this work, we propose a novel framework that eliminates the reliance on class number assumptions, enabling the model to operate without predefined task constraints. By integrating multi-level contrastive learning with a disentanglement loss, our framework captures more fine-grained features from the data. 
Our experiments reveal that CD achieves comparable performance to class-dependent methods on smaller datasets (CIFAR-10, CIFAR-100), and outperforms them on more complex datasets (STL-10, ImageNet-10), despite its simple architecture. Ablation studies further confirm the effectiveness of each component. Moreover, our framework ensures that the final model outputs are enriched with semantic information, enhancing the interpretability of the learned features.

\section{Acknowledgments}

This paper was supported by the Fujian Provincial Natural Science Foundation of China (No.2021J01129), the Fujian Provincial Higher Education Technology Research Association Fund Project (No.H2000134A), and the Fujian Agriculture and Forestry University Horizontal Technology Innovation Fund (No.KHF190015).

\appendix

\section{t-SNE Visualization of Backbone and Feature Predictor Outputs}
\label{app:tsne}
\begin{figure}[H]
    \centering
    \begin{subfigure}[b]{0.45\linewidth}
        \centering
        \includegraphics[width=\linewidth]{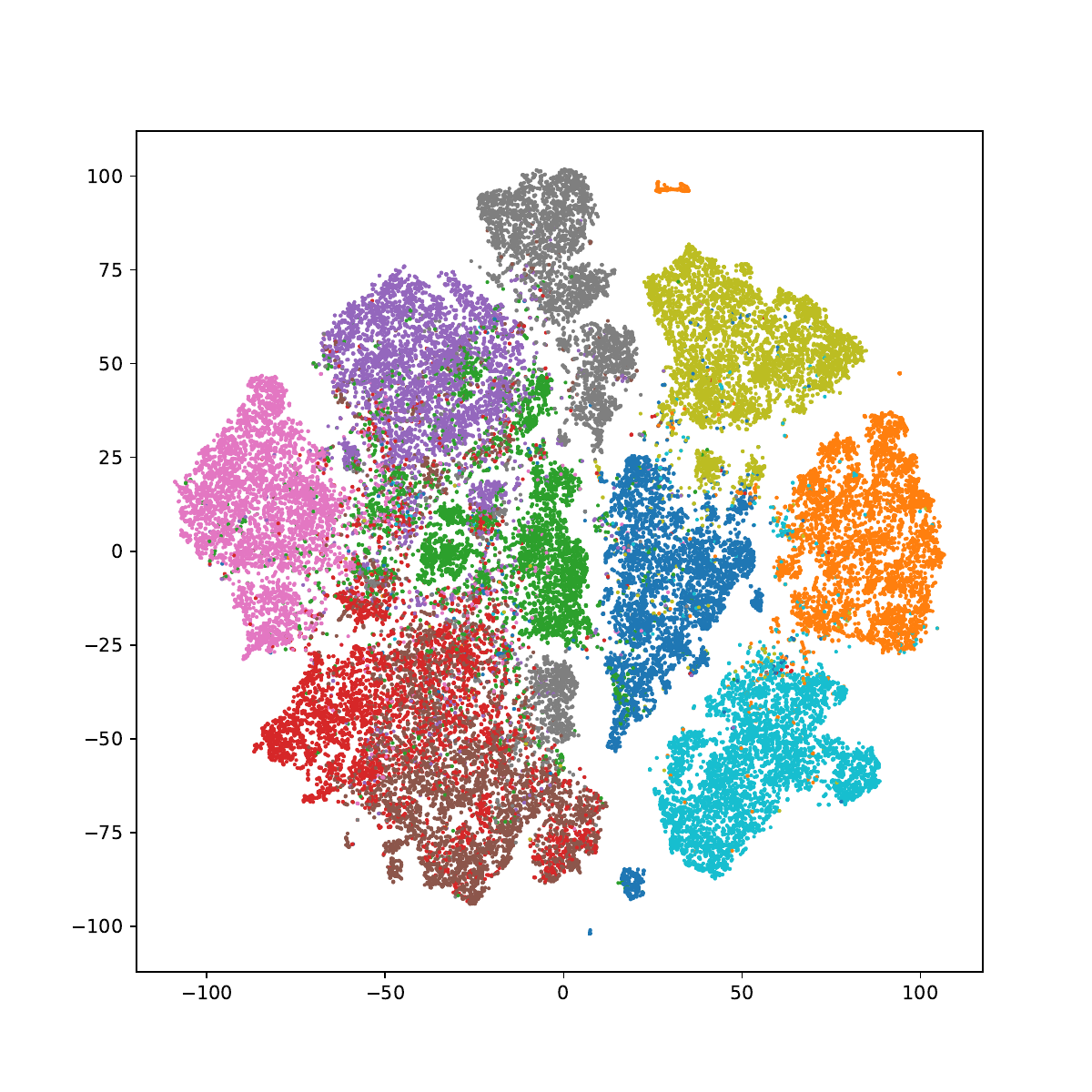}
        \caption{Backbone}
        \label{fig:tsne_backbone_cifar10}
    \end{subfigure}
    \hfill
    \begin{subfigure}[b]{0.45\linewidth}
        \centering
        \includegraphics[width=\linewidth]{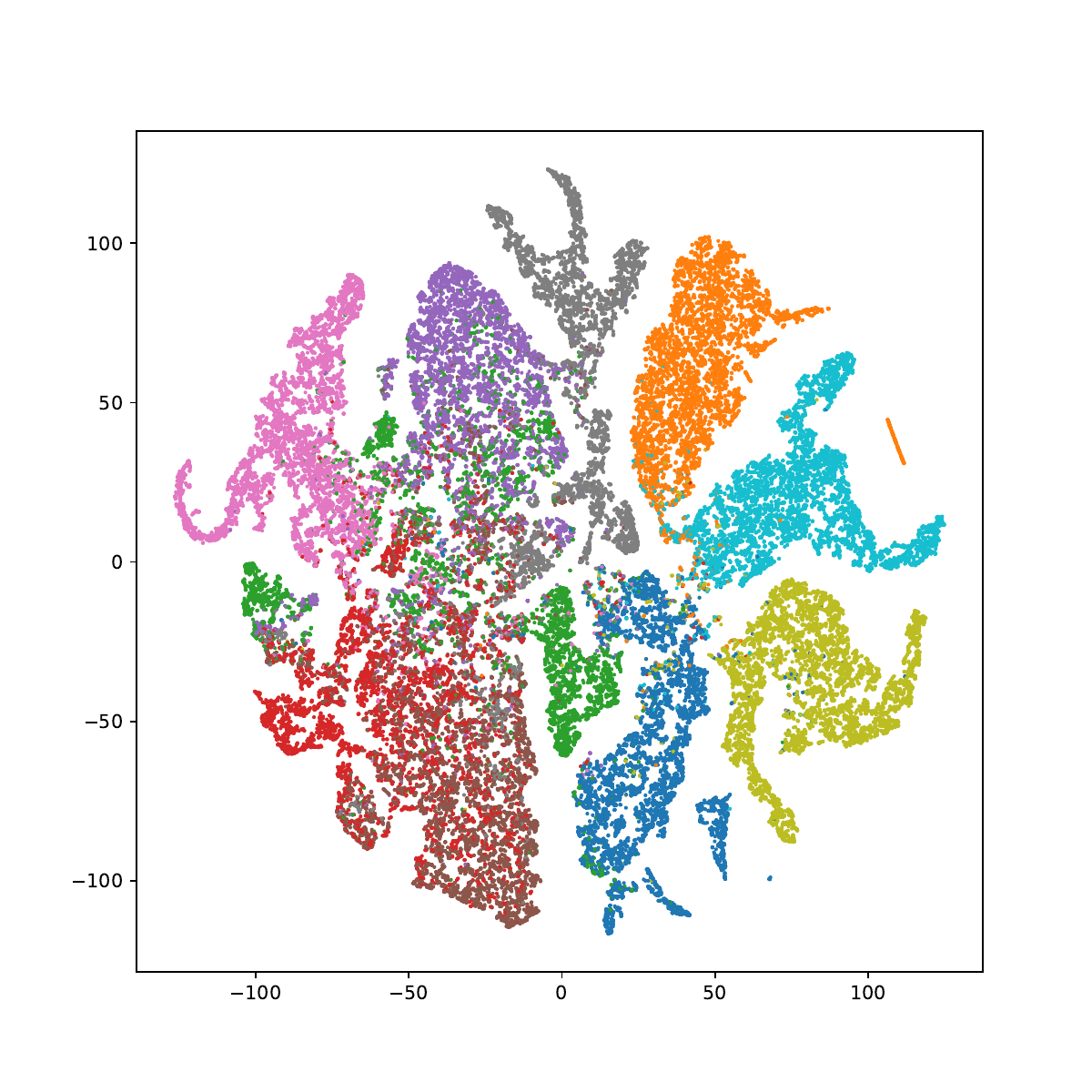}
        \caption{Feature}
        \label{fig:tsne_feature_cifar10}
    \end{subfigure}
    \caption{t-SNE visualization for CIFAR-10}
    \label{fig:tsne_cifar10}
\end{figure}

\begin{figure}[H]
    \centering
    \begin{subfigure}[b]{0.45\linewidth}
        \centering
        \includegraphics[width=\linewidth]{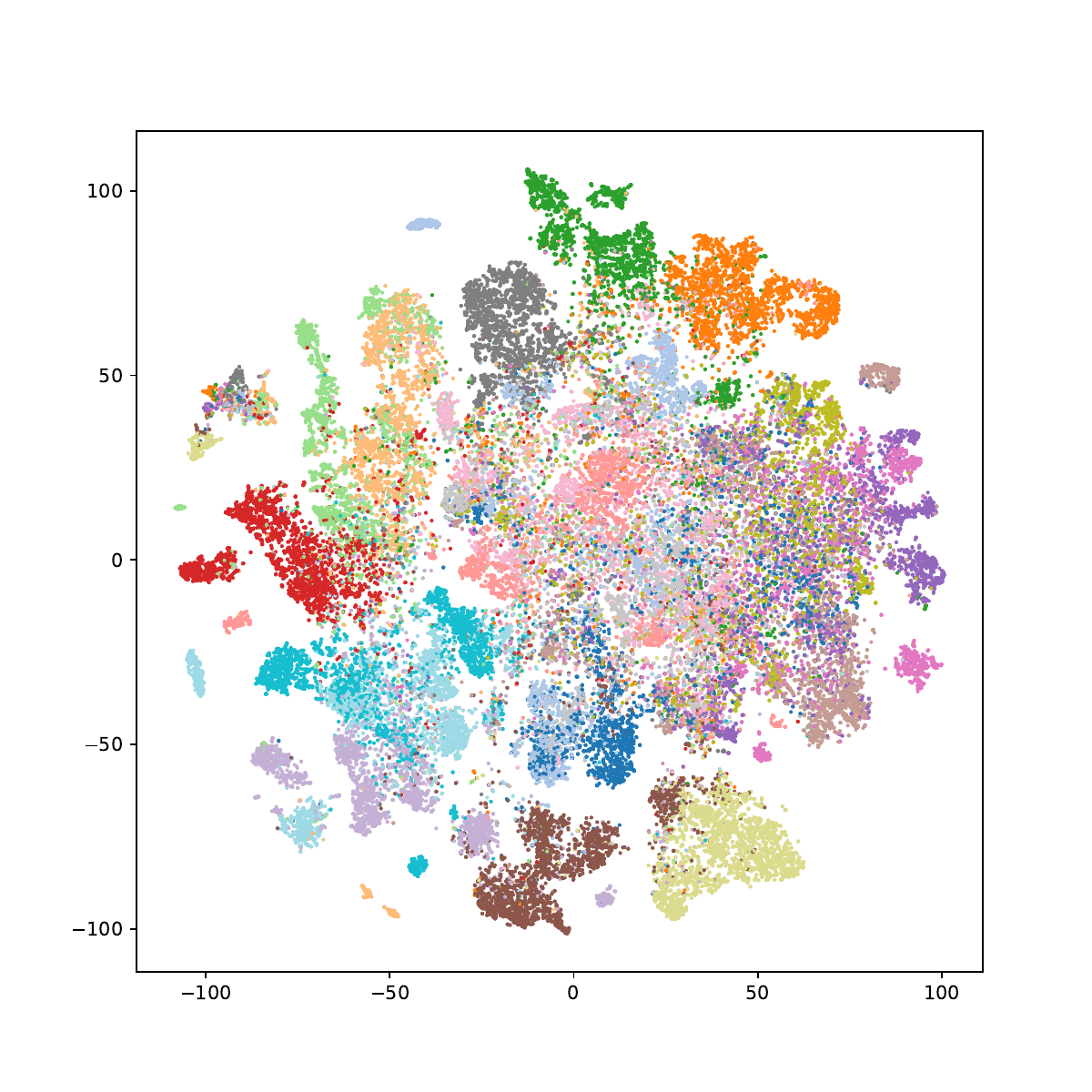}
        \caption{Backbone}
        \label{fig:tsne_backbone_cifar100}
    \end{subfigure}
    \hfill
    \begin{subfigure}[b]{0.45\linewidth}
        \centering
        \includegraphics[width=\linewidth]{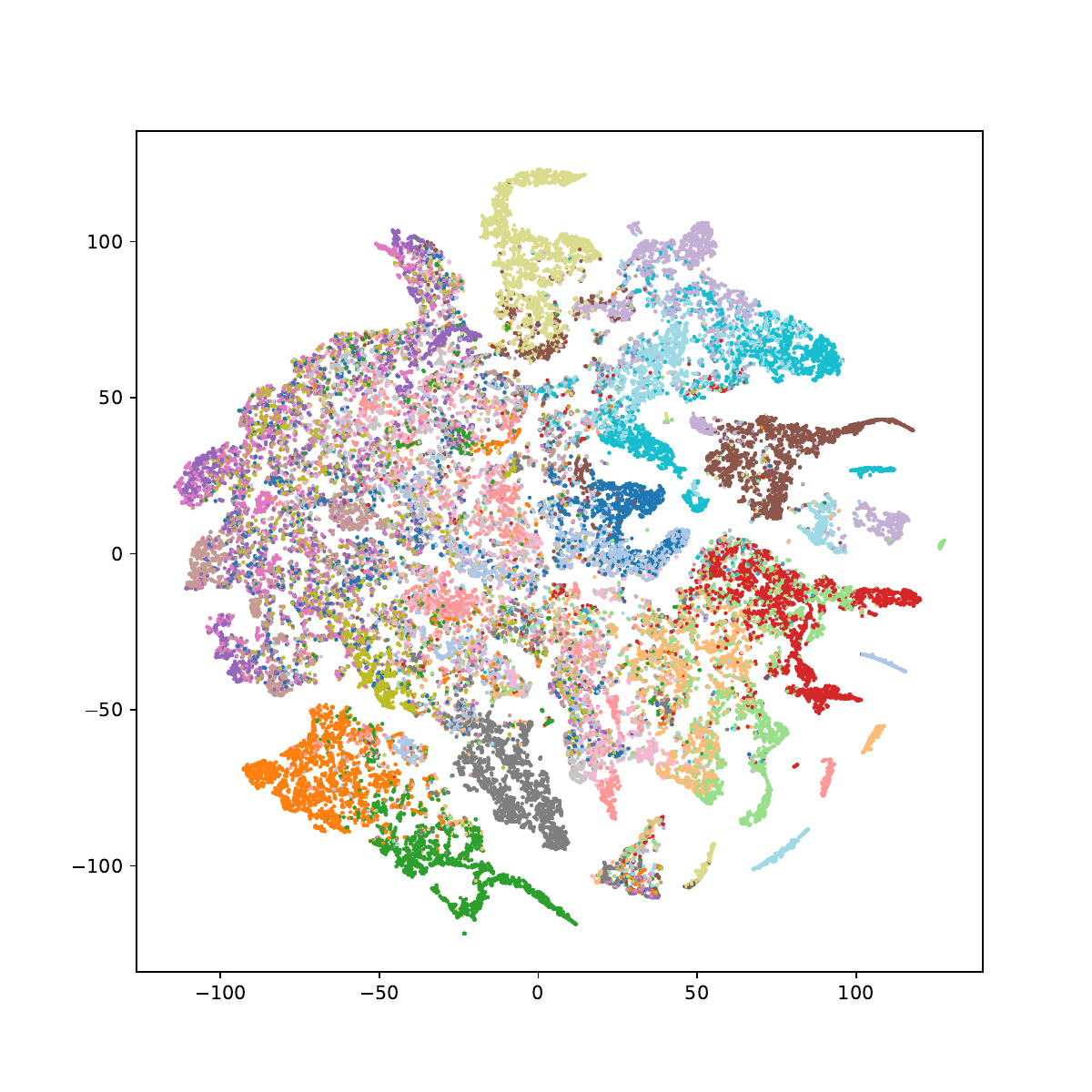}
        \caption{Feature}
        \label{fig:tsne_feature_cifar100}
    \end{subfigure}
    \caption{t-SNE visualization for CIFAR-100}
    \label{fig:tsne_cifar100}
\end{figure}

\begin{figure}[H]
    \centering
    \begin{subfigure}[b]{0.45\linewidth}
        \centering
        \includegraphics[width=\linewidth]{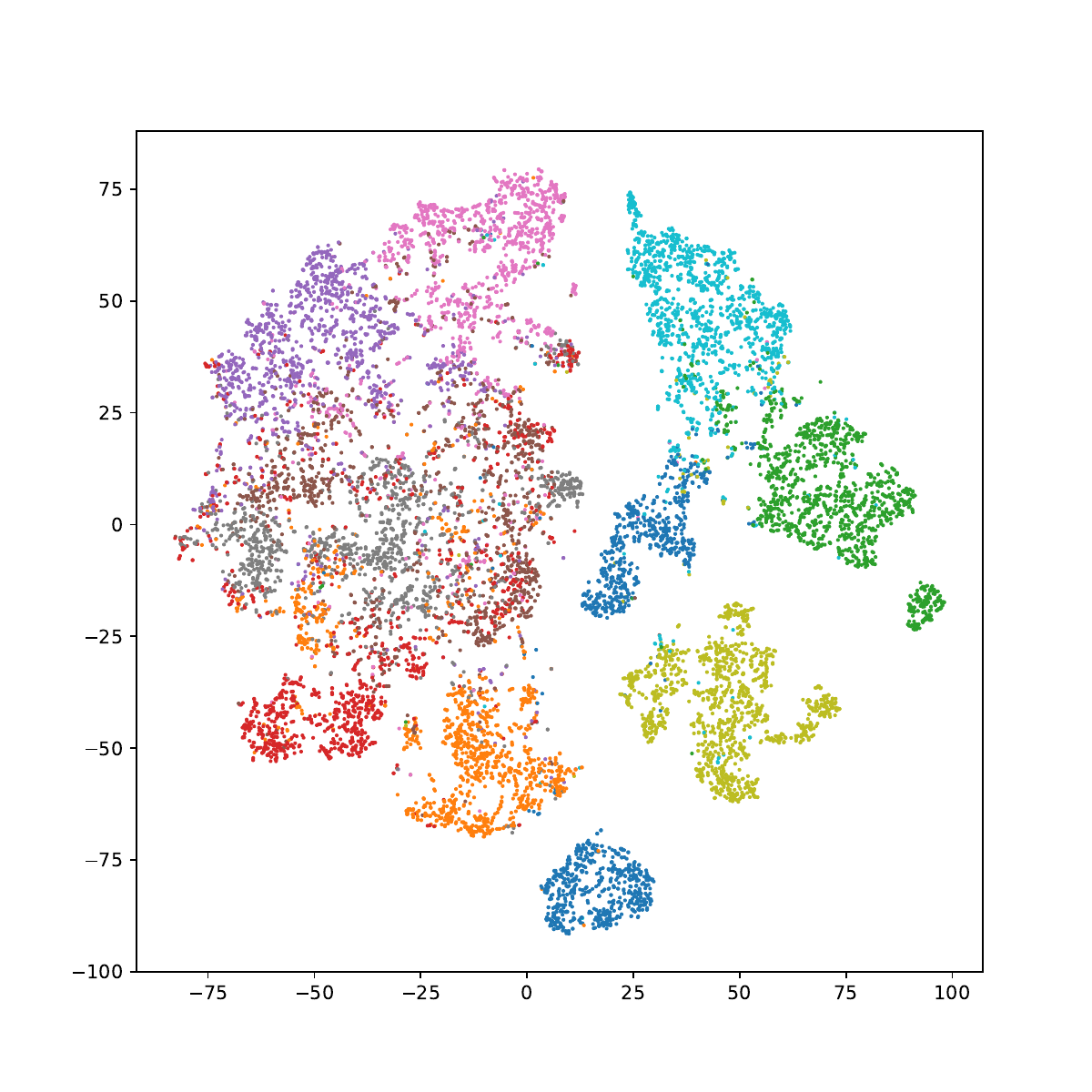}
        \caption{Backbone}
        \label{fig:tsne_backbone_stl10}
    \end{subfigure}
    \hfill
    \begin{subfigure}[b]{0.45\linewidth}
        \centering
        \includegraphics[width=\linewidth]{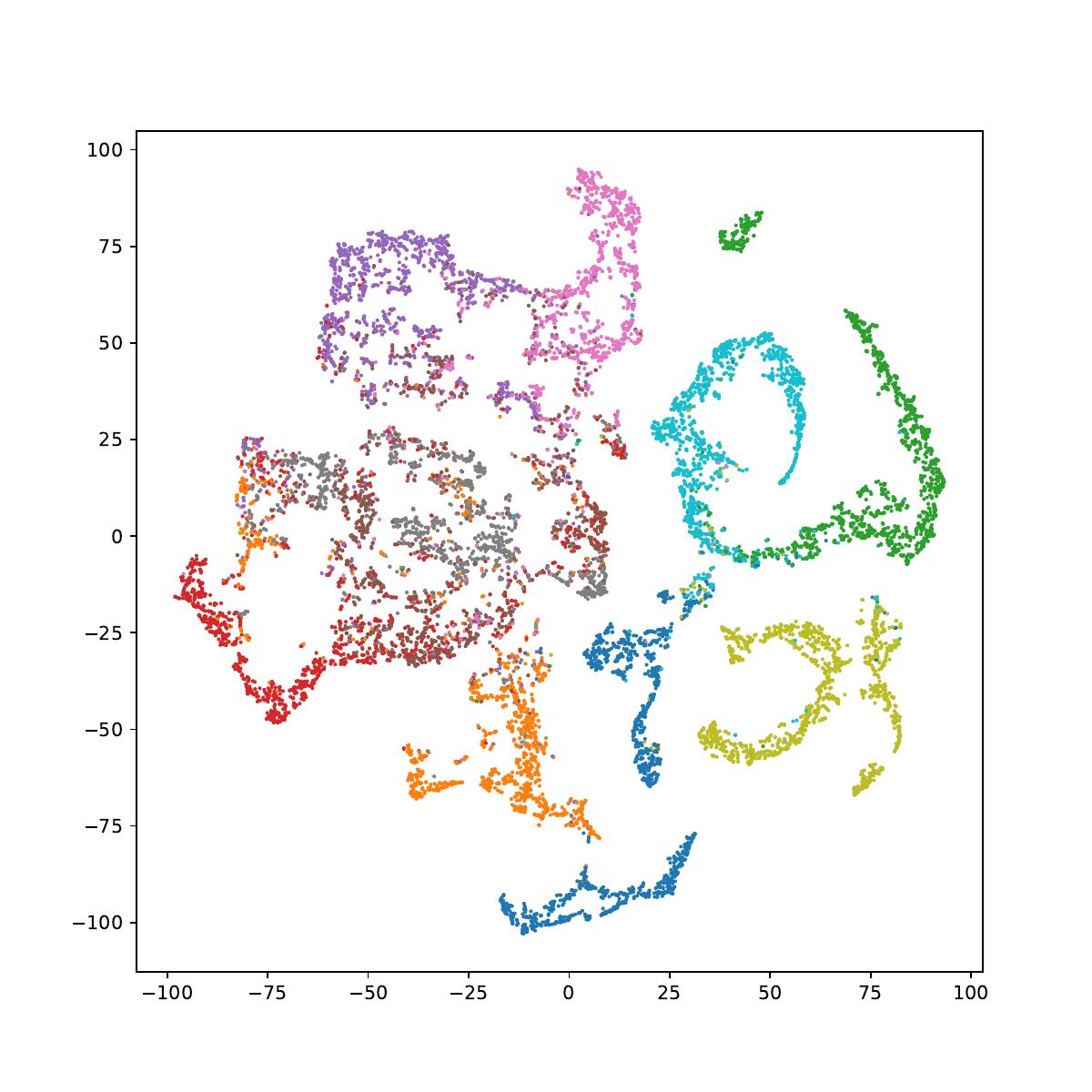}
        \caption{Feature}
        \label{fig:tsne_feature_stl10}
    \end{subfigure}
    \caption{t-SNE visualization for STL-10}
    \label{fig:tsne_stl10}
\end{figure}

\begin{figure}[H]
    \centering
    \begin{subfigure}[b]{0.45\linewidth}
        \centering
        \includegraphics[width=\linewidth]{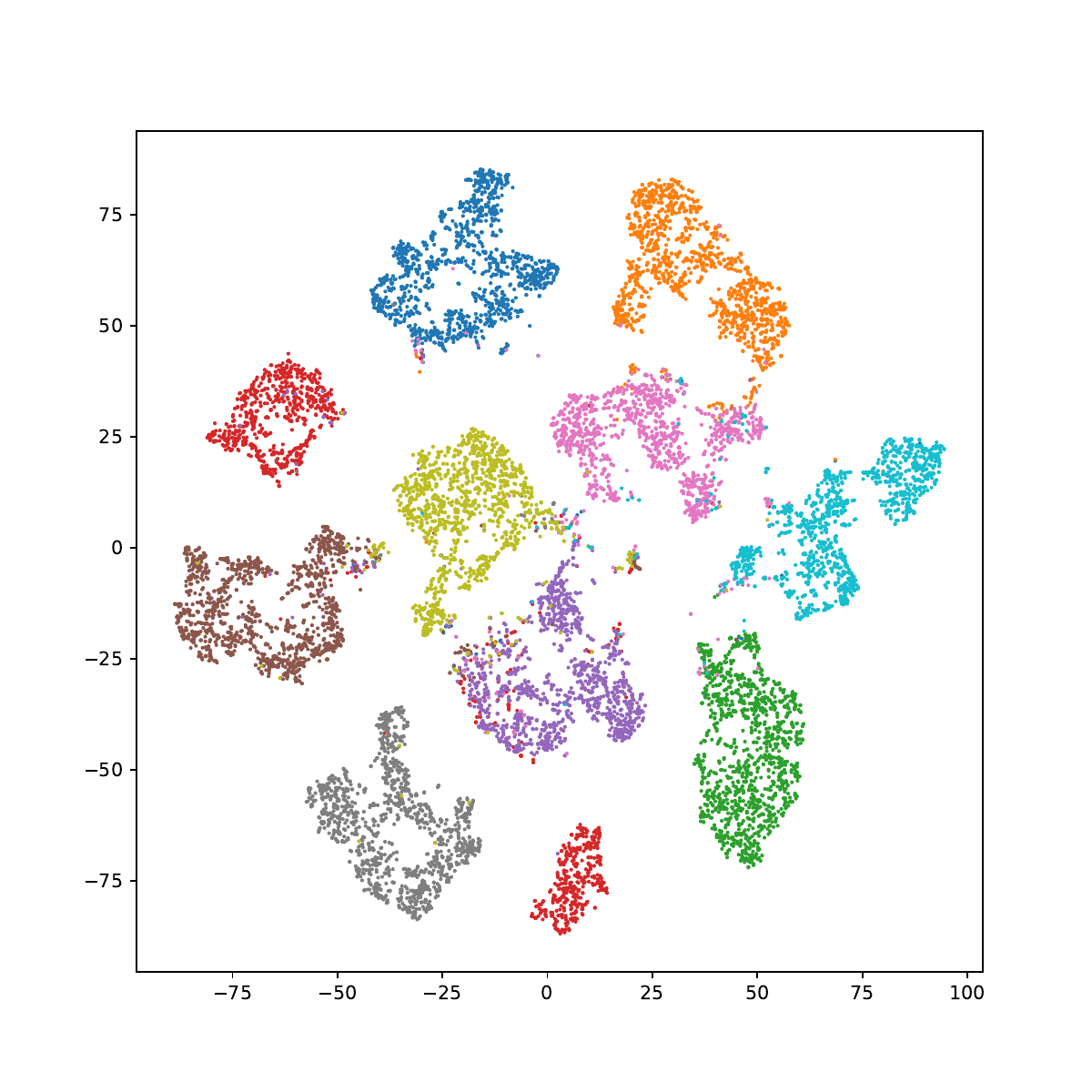}
        \caption{Backbone}
        \label{fig:tsne_backbone_imagenet10}
    \end{subfigure}
    \hfill
    \begin{subfigure}[b]{0.45\linewidth}
        \centering
        \includegraphics[width=\linewidth]{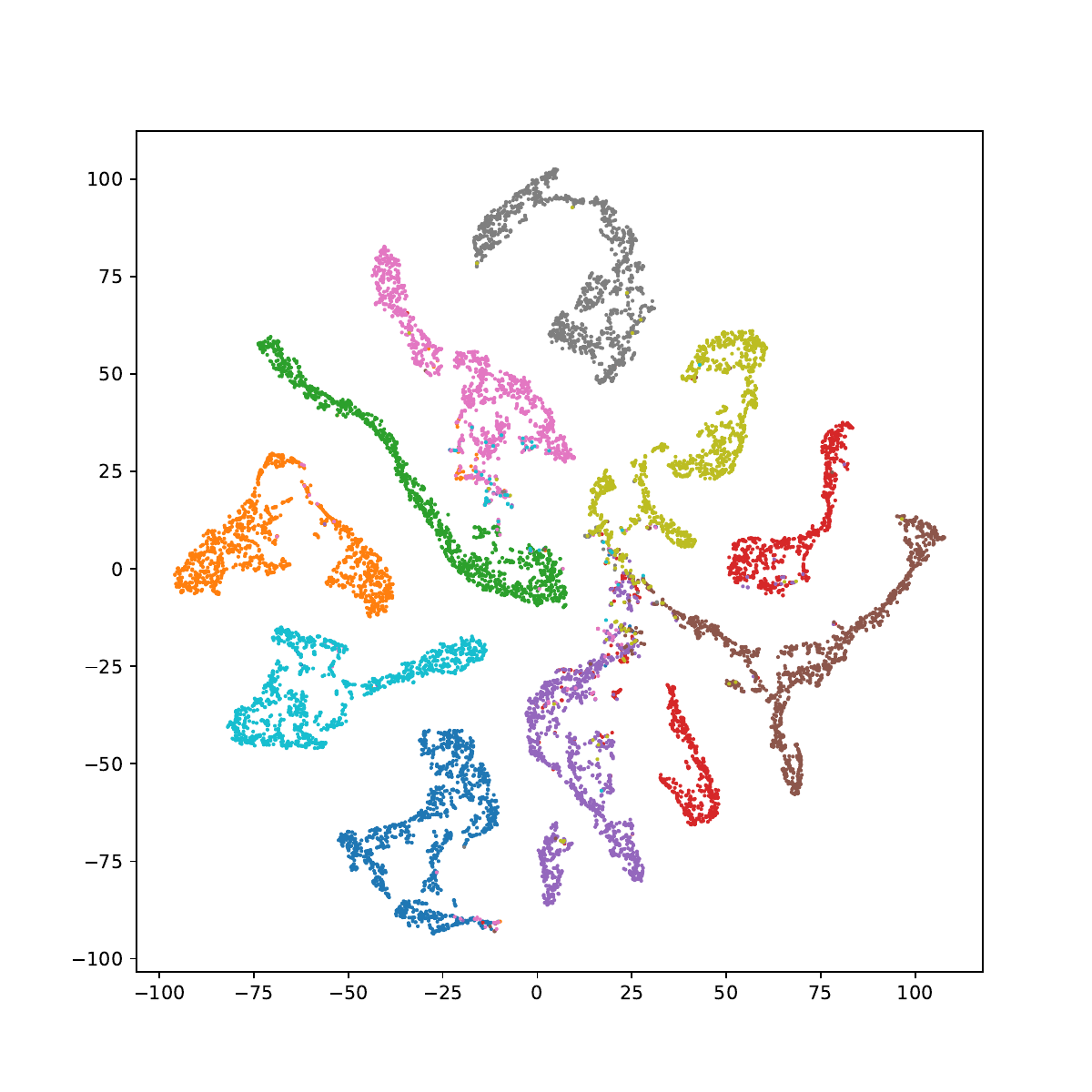}
        \caption{Feature}
        \label{fig:tsne_feature_imagenet10}
    \end{subfigure}
    \caption{t-SNE visualization for ImageNet-10}
    \label{fig:tsne_imagenet10}
\end{figure}

In this appendix, we present the t-SNE (t-Distributed Stochastic Neighbor Embedding) visualizations \cite{tsne} of the outputs both the backbone feature extraction layer and the feature predictor of our model, across all datasets: CIFAR-10, CIFAR-100, STL-10, and ImageNet-10. These visualizations are shown in Figures \ref{fig:tsne_cifar10}-\ref{fig:tsne_imagenet10}, where (a) represents the backbone output and (b) represents the feature predictor output for each dataset.

\section{Visualizing Feature Prediction Heads with LIME}
\label{app:lime}

We utilize the \textbf{LIME} (Local Interpretable Model-Agnostic Explanations) technique \cite{LIME} to visualize the outputs of the feature prediction heads. LIME is configured with $num\_samples$=5000, meaning that for each image, LIME generates 5000 perturbed samples to create an interpretable explanation. We only visualize images where the feature prediction heads produce outputs greater than \textbf{0.75}.
\begin{figure*}[ht]
\centering
\includegraphics[width=0.8\textwidth]{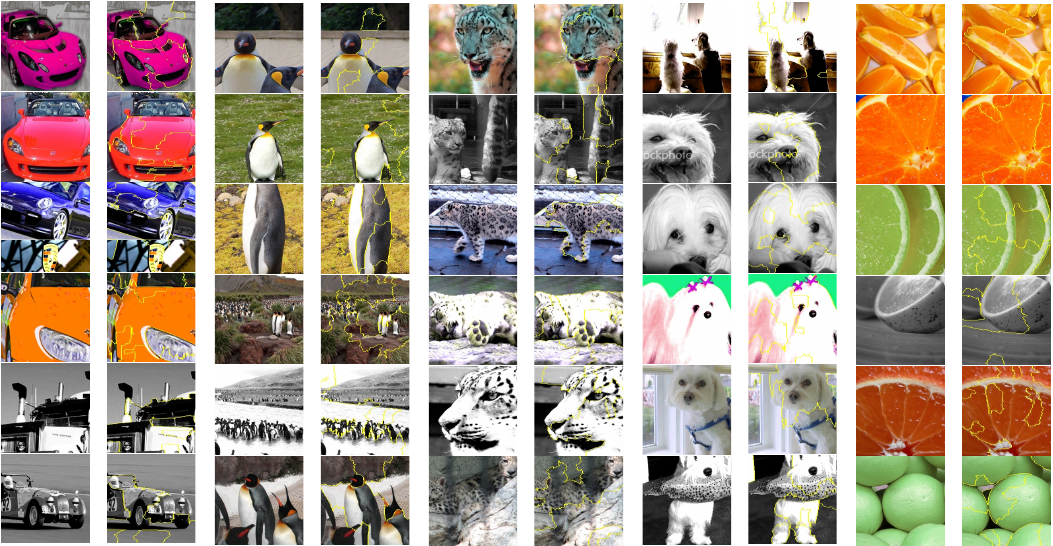}
\caption{Visualization of the attention from five feature prediction heads, with yellow-circled areas highlighting the regions that the heads focus on.}
\label{fig:lime}
\end{figure*}

The feature prediction heads visualized in Figure \ref{fig:lime} correspond to the indices 1, 5, 6, 17, and 18. These specific heads were selected to ensure a visually meaningful comparison, as some other heads produced fewer than six images that met the 0.75 threshold. The yellow highlighted areas indicate the regions that these heads attend to. 

\begin{figure*}[ht]
\centering
\includegraphics[width=0.8\textwidth]{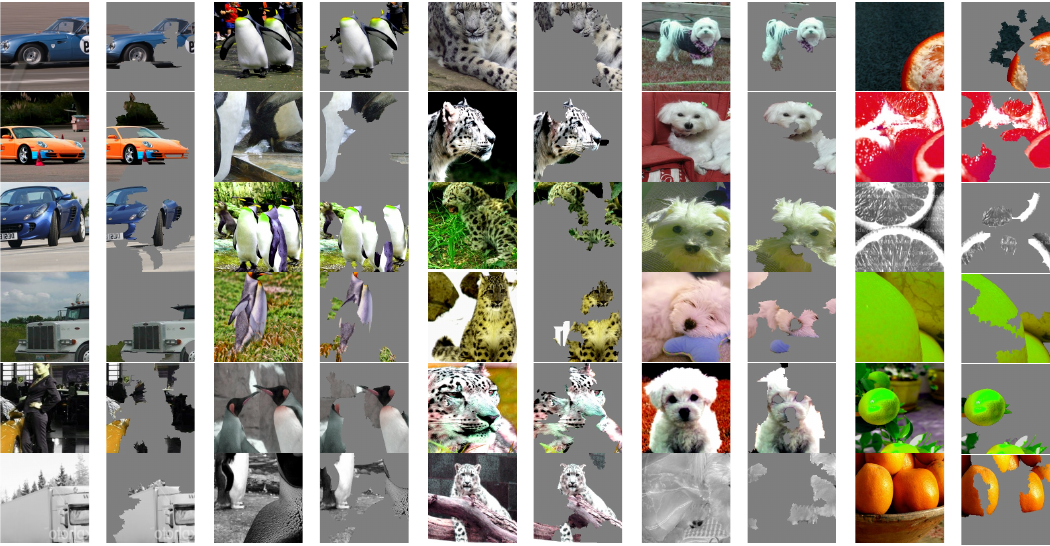}
\caption{Visualization of the same five feature prediction heads on a new batch of images, with non-attended regions masked out.}
\label{fig:lime2}
\end{figure*}

Figure \ref{fig:lime2} presents the same heads applied to a new batch of images, with a mask applied to remove areas that were not attended to.

The model used for this visualization was pre-trained on the \textbf{ImageNet-10} dataset with the following hyperparameters: batch size=256, epochs=1000, and learning rate=0.0003. All other settings were kept as defaults from our \href{https://github.com/Hoper-J/Contrastive-Disentangling/blob/master/config/config.yaml}{repository}, where the pre-trained model files and complete experimental processes for all other experiments in this study are also available.

\section{Ablation Studies on Backbone Feature Extraction}
\label{app:ablation}

This section provides additional ablation studies focused on evaluating the backbone feature extraction capabilities of our proposed framework. These experiments assess the contribution of various framework components by examining their impact on the backbone’s ability to extract meaningful features. Specifically, we analyze the performance of the backbone under different configurations by removing or modifying key components such as the instance projector, feature predictor, and data augmentation strategies.

The experiments are conducted on the STL-10 and ImageNet-10 datasets with the same batch size of 128, as used in the main experiments. We report the results using metrics such as Normalized Mutual Information (NMI), Adjusted Rand Index (ARI), and Accuracy (ACC). The backbone performance is compared across several configurations to highlight the individual and combined impact of the modules on feature extraction.

\begin{table*}[ht]
  \centering
  \caption{Effect of cosine decay schedule (Sch.) and gradient clipping (Clip.) on feature extraction.}
    \begin{tabular}{rlccc}
    \toprule
    \multicolumn{1}{l}{Dataset} & Scheduler \& & \multicolumn{3}{l}{Backbone} \\
\cmidrule{3-5}          & Grad Clipping & \multicolumn{1}{l}{NMI} & \multicolumn{1}{l}{ARI} & \multicolumn{1}{l}{ACC} \\
    \midrule
    \multicolumn{1}{l}{STL-10} & -     & $\mathbf{0.677 \pm 0.020}$ & $\mathbf{0.547 \pm 0.031}$ & $\mathbf{0.705 \pm 0.039}$ \\
          & \checkmark     & $0.674 \pm 0.015$ & $0.545 \pm 0.032$ & $0.701 \pm 0.034$ \\
    \midrule
    \multicolumn{1}{l}{ImageNet-10} & -     & $0.857 \pm 0.021$ & $0.772 \pm 0.068$ & $0.858 \pm 0.071$ \\
          & \checkmark     & $\mathbf{0.886 \pm 0.008}$ & $\mathbf{0.844 \pm 0.022}$ & $\mathbf{0.918 \pm 0.026}$ \\
    \bottomrule
    \end{tabular}%
  \label{tab:appendix_sch_clip}%
\end{table*}
\begin{table*}[ht]
  \centering
  \caption{Effect of feature-level head on feature extraction.}
    \begin{tabular}{rlccc}
    \toprule
    \multicolumn{1}{l}{Dataset} & Feature-Level & \multicolumn{3}{l}{Backbone} \\
\cmidrule{3-5}          &       & \multicolumn{1}{l}{NMI} & \multicolumn{1}{l}{ARI} & \multicolumn{1}{l}{ACC} \\
    \midrule
    \multicolumn{1}{l}{STL-10} & -     & $0.659 \pm 0.016$ & $0.530 \pm 0.028$ & $0.692 \pm 0.030$ \\
          & \checkmark     & $\mathbf{0.674 \pm 0.015}$ & $\mathbf{0.545 \pm 0.032}$ & $\mathbf{0.701 \pm 0.034}$ \\
    \midrule
    \multicolumn{1}{l}{ImageNet-10} & -     & $0.870 \pm 0.022$ & $0.809 \pm 0.064$ & $0.886 \pm 0.058$ \\
          & \checkmark     & $\mathbf{0.886 \pm 0.008}$ & $\mathbf{0.844 \pm 0.022}$ & $\mathbf{0.918 \pm 0.026}$ \\
    \bottomrule
    \end{tabular}%
  \label{tab:appendix_featurelevel}%
\end{table*}%
\begin{table}[ht]
  \centering
  \caption{Effect of single-view vs. dual-view data augmentation on feature extraction.}
    \begin{tabular}{rllccc}
    \toprule
    \multicolumn{1}{l}{Dataset} & \multicolumn{2}{l}{Augmentation} & \multicolumn{3}{l}{Backbone} \\
\cmidrule{2-6}          & x1    & x2    & \multicolumn{1}{l}{NMI} & \multicolumn{1}{l}{ARI} & \multicolumn{1}{l}{ACC} \\
    \midrule
    \multicolumn{1}{l}{STL-10} & -     & \checkmark     & $0.660 \pm 0.011$ & $0.520 \pm 0.022$ & $0.683 \pm 0.022$ \\
          & \checkmark     & \checkmark     & $\mathbf{0.671 \pm 0.019}$ & $\mathbf{0.545 \pm 0.032}$ & $\mathbf{0.701 \pm 0.034}$ \\
    \midrule
    \multicolumn{1}{l}{ImageNet-10} & -     & \checkmark     & $\mathbf{0.889 \pm 0.010}$ & $\mathbf{0.851 \pm 0.024}$ & $\mathbf{0.923 \pm 0.018}$ \\
          & \checkmark     & \checkmark     & $0.886 \pm 0.008$ & $0.844 \pm 0.022$ & $0.918 \pm 0.026$ \\
    \bottomrule
    \end{tabular}%
  \label{tab:appendix_augmentation}%
\end{table}

It can be observed that when evaluating the model’s ability as a pretext model, or more specifically, the feature extraction capability of the backbone, the scheduler, gradient clipping, and dual-view data augmentation are not essential. However, if we aim for the final output of the model to be meaningful, rather than focusing solely on intermediate states, these components become crucial. You may refer to Table \ref{tab:sch&clip} - Table \ref{tab:augmentation}, especially Table \ref{tab:feature-level}, where our model demonstrates significantly higher semantic expressiveness in the final layer.

\bibliographystyle{elsarticle-num-names} 
\bibliography{refs}

\end{document}